\documentclass[journal]{IEEEtran}

\usepackage{cite}

\usepackage{color}

\ifCLASSINFOpdf
 \usepackage[pdftex]{graphicx}
\else
 \fi
\usepackage{amsmath}
\usepackage{algorithm}
\usepackage{algpseudocode}
\usepackage{array}
\usepackage{fixltx2e}
\usepackage{stfloats}
\usepackage{url}
\usepackage{comment}
\usepackage{xcolor}
\usepackage{subcaption}

\begin{document}
% \title{Federated Testing for Federating Learning (FedTest): A Promising Scheme to Improve the Convergence Rate and Mitigate the Effects of Adversarial Attacks}

\title{Federated Testing (FedTest): A New Scheme to Enhance Convergence and Mitigate Adversarial Attacks in Federating Learning}


\author{
\IEEEauthorblockN{Mustafa Ghaleb,  Mohanad Obeed,~\IEEEmembership{Member,~IEEE,} Muhamad Felemban,~\IEEEmembership{Senior Member,~IEEE}, Anas Chaaban,~\IEEEmembership{Senior Member,~IEEE},
Halim Yanikomeroglu~\IEEEmembership{Fellow,~IEEE}}

% \author{\IEEEauthorblockN{Michael Shell\IEEEauthorrefmark{1},
% Homer Simpson\IEEEauthorrefmark{2},
% James Kirk\IEEEauthorrefmark{3}, 
% Montgomery Scott\IEEEauthorrefmark{3}, and
% Eldon Tyrell\IEEEauthorrefmark{4},~\IEEEmembership{Fellow,~IEEE}}
% \IEEEauthorblockA{\IEEEauthorrefmark{1}School of Electrical and Computer Engineering,
% Georgia Institute of Technology, Atlanta, GA 30332 USA}
% \IEEEauthorblockA{\IEEEauthorrefmark{2}Twentieth Century Fox, Springfield, USA}
% \IEEEauthorblockA{\IEEEauthorrefmark{3}Starfleet Academy, San Francisco, CA 96678 USA}
% \IEEEauthorblockA{\IEEEauthorrefmark{4}Tyrell Inc., 123 Replicant Street, Los Angeles, CA 90210 USA}% <-this % stops an unwanted space
\thanks{M. Ghaleb is with the Center for Intelligent Secure Systems,  King Fahd University of Petroleum and Minerals (KFUPM), Dhahran 31261, Saudi Arabia. M. Obeed and H. Yanikomeroglu are with the Systems and Computer Engineering Department, Carleton University, Ottawa, ON K1S 5B6, Canada. M. Felemban is with the Computer Engineering Department and Center for Intelligent Secure Systems,  KFUPM, Dhahran 31261, Saudi Arabia. A. Chaaban is with the School of Engineering, University of British Columbia, Kelowna, BC V1V1V7, Canada.\\
}}

% \markboth{Journal of \LaTeX\ Class Files,~Vol.~14, No.~8, August~2015}%
% {Shell \MakeLowercase{\textit{et al.}}: Bare Demo of IEEEtran.cls for IEEE Transactions on Magnetics Journals}



% for Transactions on Magnetics papers, we must declare the abstract and
% index terms PRIOR to the title within the \IEEEtitleabstractindextext
% IEEEtran command as these need to go into the title area created by
\IEEEtitleabstractindextext{%
\begin{abstract}

Federated Learning (FL) has emerged as a significant paradigm for training machine learning models. This is due to its data-privacy-preserving property and its efficient exploitation of distributed computational resources. This is achieved by conducting the training process in parallel at distributed users. However, traditional FL strategies grapple with difficulties in evaluating the quality of received models, handling unbalanced models, and reducing the impact of detrimental models. To resolve these problems, we introduce a novel federated learning framework, which we call federated testing for federated learning (FedTest). In the FedTest method, the local data of a specific user is used to train the model of that user and test the models of the other users.  This approach enables users to test each other's models and determine an accurate score for each. This score can then be used to aggregate the models efficiently and identify any malicious ones.  Our numerical results reveal that the proposed method not only accelerates convergence rates but also diminishes the potential influence of malicious users. This significantly enhances the overall efficiency and robustness of FL systems.

\end{abstract}

}

\maketitle

\IEEEdisplaynontitleabstractindextext
\IEEEpeerreviewmaketitle

\section{Introduction}

%{\color{red}{Please start by motivating machine learning, then mention the centralized machine learning drawbacks like their bandwidth consumption and data privacy.}}
%1- Motivating ML
%2- Introducing centralized ML and mentioning its drawbacks (privacy concern, BW consumption, Edge devices resources exploitation, edge computing)
%3- Defining FL and discuss its advantages over the centralized one
%4- Literature review,
%5- Discussing the limitation of the existing work (detecting the malicious attacks and evaluating the received models)
%6- Introduce the proposed FL scheme and show how it solves the above issues
%7- Paper organizaiton

The growing interest in machine learning (ML) stems from the fact that it provides promising results in various fields, including healthcare, finance, and technology. The power of ML approaches comes from the unprecedented availability of data.   In centralized ML, the datasets must be collocated in one device to train an ML model efficiently. The availability of datasets on one device guarantees fast convergence in the training stage and high-accuracy models in the testing stage.  However, in many applications, datasets are distributed over users' devices, which are often not willing to share. Even if users are willing to share their data with a centralized server, the massive amount of data introduces a challenge in transferring them to a server due to the scarcity of communication resources. Fortunately, the advances in storage and computation capabilities in users' devices open an opportunity for training ML models on these devices in a distributed manner. This attracts researchers and engineers to investigate distributed ML methods to leverage distributed resources, preserve user privacy, and minimize communication cost. A common distributed ML method is called federated learning (FL).

%The FL server aggregates the models based on 

\begin{figure}
    \centering
    \includegraphics[width=0.5\textwidth]{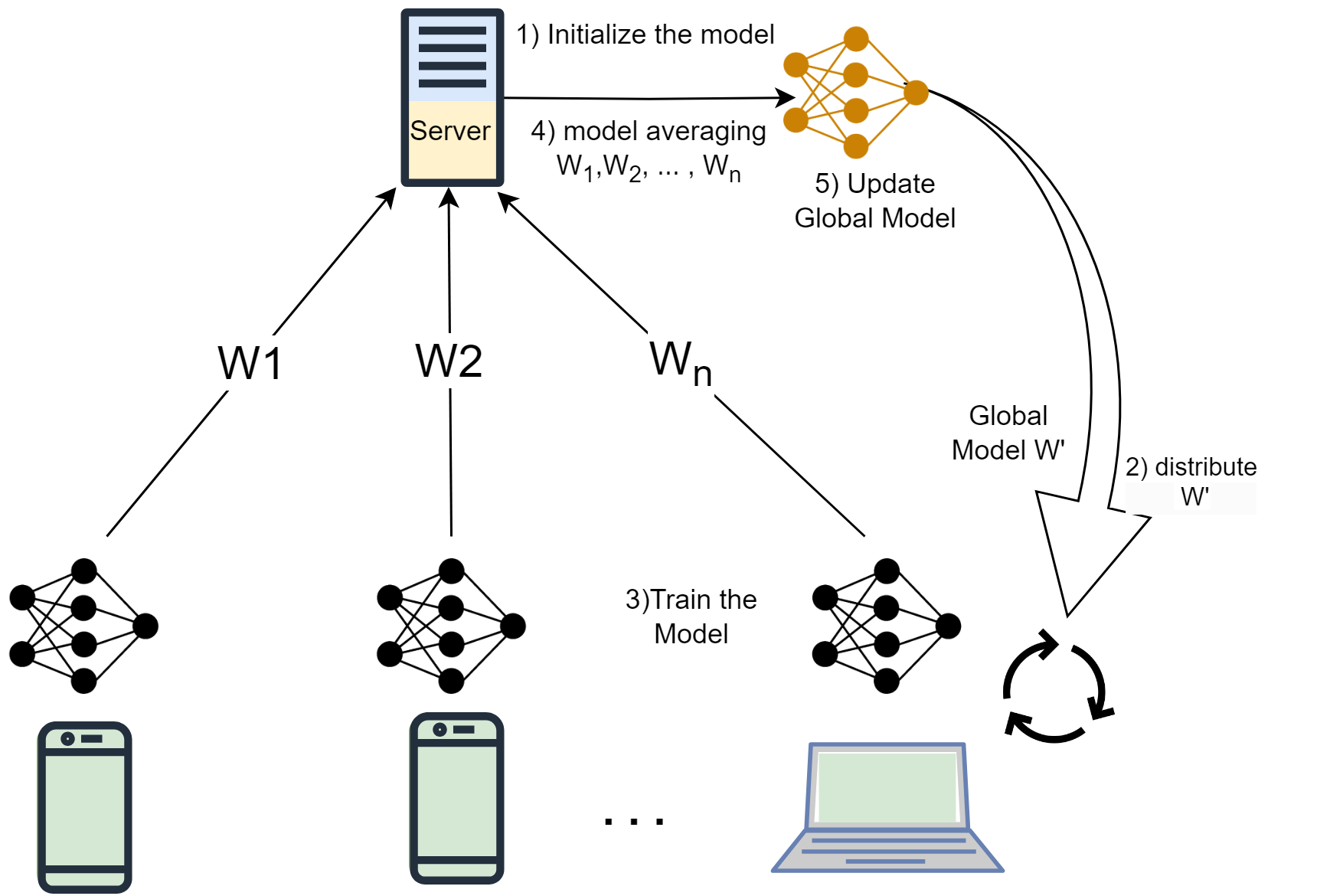}
    \caption{FedAvg method where the models are aggregated based on the dataset sizes.}
    \label{FedAvg}
\end{figure}

There are several approaches to aggregate local models on an FL server. The most common approach is federated averaging (FedAvg) \cite{mcmahan2017communication}.  In the Fedavg approach, the FL server aggregates the models by averaging them and weighting each model based on the number of its samples. Then it broadcasts the updated global model back to the devices involved for the next round of training. Figure \ref{FedAvg} shows the FedAvg method where the models are aggregated based on the dataset sizes.

Despite its groundbreaking contributions to decentralized machine learning, FedAvg alone does not completely address the intricate complexities of the federated learning landscape. As we explore this area, we encounter several additional hurdles that demand innovative solutions. Basic federated learning approaches face challenges related to fast convergence, robustness against unbalanced and non-IID data distribution, and resilience against malicious clients. In particular, the global model in traditional FL approaches may diverge in realistic situations, mainly when data are non-identically distributed across devices and/or significant variations in dataset sizes exist. In addition, in some cases, the quality of the data may vary with the users, which traditional FL approaches do not consider in their aggregation procedure. In particular, weighting models according to the number of samples is not efficient, since the value of each model must be based on how much it contributes to improving the model performance. For example, it is not accurate to weight a user's model which was trained on thousands of samples from two classes higher than a model from another user that was trained on hundreds of samples from all classes \cite{chai2020tifl}.

In addition to the foundational principles and challenges discussed so far, it is also important to consider the inherent risk associated with FL adversarial attacks. These malicious attempts are engineered to impair the performance of the model. FL's unique structure, leveraging a vast network of untrusted devices and private, non-inspectable datasets for its training process, is markedly different from traditional approaches. However, this characteristic potentially opens up novel attack surfaces during the training phase, which amplifies the risk factor.
%, distinguishing them from data inference attacks that illicitly mine confidential information from users' private data. 
Adversarial attacks can take various forms, such as model evasion attacks, model update poisoning, and data poisoning. Typically, these attacks are classified according to the stage of their implementation: poisoning attacks occur during the training phase, while evasion attacks occur during the inference phase. As we continue to dig deeper into the domain of FL and confront its multifaceted challenges, it is crucial to keep a vigilant eye on these potential security threats and develop robust countermeasures. For example, one challenge in FedAvg is that it cannot distinguish harmful models from trusted ones. In many cases, some devices may have malicious tendencies and start to share harmful models aiming at slowing down the convergence rate or contaminating the global model. In such cases, FedAvg cannot handle these malicious behaviors and would fail to achieve the required accuracy rate.

One way to address these challenges is to implement a fast-convergent federated learning approach that uses test data for model aggregation \cite{chai2020tifl}. This method automatically adjusts the weights of the client models based on their accuracy in the server-hosted testing data, resulting in an accuracy-based process for updating the weights of the generalized server model. Notably, this approach exhibits robustness against unbalanced and non-IID data distribution compared to the FedAvg method. In addition, it demonstrates resilience against malicious clients. However, the accuracy performance of the received models is difficult to obtain because of the unavailability of testing data at the federated server that should span over all classes. Using such testing data for training rather than for testing would be more beneficial if such data is available.

This paper proposes a new FL framework (FedTest) to solve the aforementioned issues and improve the convergence rate. \textit{FedTest} leverages user-specific data for local model training and evaluation of other user models, thus fully utilizing the distributed nature of FL for both training and testing while ensuring data privacy. This novel technique accelerates convergence rates, mitigates the potential influence of malicious users, and improves the overall efficacy and robustness of FL systems. We demonstrate its potential to effectively counter FL's challenges and contribute significantly to the evolution of this domain.

%Conventional Federated Learning strategies encounter obstacles in determining the quality of acquired models, addressing uneven model contributions, and reducing the impact of harmful models. To tackle these concerns, Accuracy-based Federated Learning has been suggested as a solution; however, its practical implementation is restricted by the server's capacity to access ample testing datasets. Our method, Federated Testing for Federated Learning (FedTest), takes advantage of user-generated data to train the model locally and evaluate models from other users. This novel method promotes faster convergence rates and mitigates the influence of malicious users, thereby improving the overall efficacy and robustness of FL systems.

The key contributions of this paper are as follows.
\begin{enumerate}
    \item 
We propose a novel approach called \textit{FedTest}, which addresses the challenges of assessing model quality, handling unbalanced models, and mitigating the impact of harmful models in traditional Federated Learning systems.
   \item 
We show that \textit{FedTest} overcomes the limitations of accuracy-based FL by enabling users to train local models and test other users' models using their own data, thereby eliminating the server's reliance on an extensive testing dataset.
   \item 
Our approach demonstrates faster convergence rates, which contributes to improving the overall efficiency of FL systems.
   \item 
The \textit{FedTest} method successfully mitigates the influence of malicious users, thereby enhancing the robustness of FL systems and ensuring a more secure and reliable learning environment. Even with a less robust dataset like MNIST, our proposed approach demonstrates improvement in the presence of malicious users within the network.
   \item 
We comprehensively analyze and evaluate the proposed \textit{FedTest} approach, highlighting its advantages over existing methods and showcasing its potential for practical implementation in various real-world scenarios.
\end{enumerate}

It is important to note that the proposed algorithm is also based on device-to-device (D2D) communication technology. In D2D communication, the UEs can exchange information without the help of a base station~(BS) or an access point~(AP).     

\section{Traditional Federated Learning}
% An FL aims to find an optimal global learning model using the distributed dataset over the users without sharing this dataset. 
The most common approach in FL is Fedavg, where users compute a model update locally and only share these updates with a central server. Then, this server aggregates the updates, averages them based on the number of samples of each user, and sends the updated global model back to the involved users for the next round of training. This iterative process continues until convergence.

%Non-i.i.d. (non-independent and identically distributed) data samples cause inconsistencies between global and local objectives, which slows down the convergence of the FL model.

% However, FL is vulnerable to various problems, including 1) systems heterogeneity between clients that results in straggler problems and 2) statistical heterogeneity between clients that results in an issue with objective inconsistency. These issues could result in a considerable slowdown in the convergence speed of the model in the considered heterogeneous environment. Therefore, many suggested solutions are proposed to enhance the speed of model convergence. 
FedAvg and its variants \cite{konevcny2016federated,mcmahan2017communication,kairouz2021advances} are typical examples of traditional federated optimization techniques employing local model updates, where clients update their models on local datasets using multiple training iterations before communicating their models with the server. As a result, the communication needed to enable the FL model is significantly reduced. A common pitfall, termed as the \textit{client drift issue}, emerges within these traditional federated optimization techniques where client models start to diverge or \textit{drift} from the global optimum due to multiple iterations of training performed on local data. As clients continuously update their models with their own local datasets, these models may start to overfit to their specific data, resulting in a model that's optimal for their local data but far from the global optimum. This drifting of client models can slow down the overall convergence of the FL training process.

Another limitation of traditional FL algorithms emerges when we consider the quality and diversity of data held by individual users. In real-world scenarios, it is common for users to have imbalanced datasets skewed toward certain classes. For instance, in the context of a health dataset, a user's data might be heavily weighted towards a specific condition or demographic, failing to represent the complete spectrum of health conditions or populations. This lack of comprehensive representation can lead to biased learning and potentially impact the effectiveness and generalizability of the global model. Therefore, this imbalance poses a significant challenge to applying such algorithms in FL, underscoring the need for strategies that can effectively handle and learn from such imbalanced data distributions. 

Another potential issue with the traditional FedAvg approach is that some participating nodes may have malicious intentions, and they can send poisoned model parameters to the central server to affect the global model's performance. For instance, a node may deliberately send gradients that increase the loss function, resulting in a worse global model. To counter adversarial attacks, one solution is to introduce an accuracy-based weighted aggregation scheme \cite{chai2020tifl}, replacing the traditional sample-based weight in the FedAvg algorithm. This strategy operates on the premise of ascertaining each local model's performance on a held-out test set. The central server then aggregates these models using a weighted average, with weights proportional to each model's test-set accuracy. This modification privileges more accurate local models, assigning them greater weights, while less accurate or potentially malicious models receive lesser weights. As such, the deleterious influence of malicious gradients on the global model is significantly curtailed. Unlike the traditional FedAvg method, which treats all models uniformly irrespective of their test-set performance, this accuracy-centric approach promotes superior global model performance by favoring the most accurate local models. In essence, this method enhances the robustness and performance of FL under potential malicious attacks, endorsing the most accurate models and limiting the detrimental effects of less reliable ones.

In recent research, there has been a focus on adapting centralized optimization algorithms to federated learning environments to improve convergence speed. Existing algorithms, like FedAvg, perform multiple stochastic gradient descent (SGD) updates on individual clients before communication, but this can lead to overfitting to local data, resulting in slower convergence and increased communication. To overcome these limitations, new algorithms such as those proposed by Wang et al. \cite{wang2021local}, and Karimireddy et al. \cite{karimireddy2020mime} aim to adapt arbitrary centralized optimization algorithms, such as SGD with momentum or Adam, for use in federated learning settings while mitigating client drift. 
Authors in \cite{chen2020wireless} introduced Collaborative Federated Learning (CFL) to address traditional FL and centralized learning limitations. FL, which requires all devices to connect directly to a central controller, is not always feasible in large-scale wireless networks like IoT systems due to energy constraints or high transmission delays. Centralized learning can also compromise privacy. CFL allows more devices to participate in learning without needing a direct connection to a central controller.

Traditional FL approaches often involve selecting a fixed subset of clients for each training round, which may not be optimal due to the heterogeneity of client data and capabilities. Federated Adaptive Training (FedAda) \cite{zhang2022fedada} aims to address this limitation by dynamically adapting the client selection process. It leverages local training progress and performance metrics from previous rounds to estimate the potential contributions of each client. Based on these estimates, FedAda adaptively selects a subset of clients with higher potential for the current round, which leads to reduce runtime delays between clients and increase the convergence rate.

\section{FedTest}
Accuracy-based federated learning methodologies pose certain challenges, particularly concerning the effective use of datasets. A pivotal issue arises when trying to sustain a dataset on a single node or on the server that is representative of all classes. 
If the server possesses these comprehensive data samples, an intuitive solution might be to use this information for training rather than testing, which would substantially improve the convergence rate.
% However, this solution isn't entirely straightforward. 
%Allocating this rich, representative dataset solely for testing may not be the most efficient approach. Instead, leveraging such a valuable dataset for model training could significantly optimize the learning process and hasten convergence. 
This paradigm shift prompts us to rethink conventional methodologies and highlights the need for innovative strategies to exploit the full potential of datasets within federated learning frameworks.

% In our setup, the users' datasets are used for training and testing. This allows for efficient usage, as users can use their dataset to train their model and test others. However, this presents a problem in the accuracy-based approach, as it's difficult to ensure all classes are represented at the server level. If all classes are indeed present, using this data for training is often more beneficial than testing, enhancing the training process's efficiency. Moreover, this setup breaks down if we operate under the assumption of a non-IID dataset, a common situation in real-world applications of federated learning. With a non-IID dataset, the assumption that all classes can be present on the server for accuracy-based testing becomes unrealistic. This adds a layer of complexity to applying an accuracy-based approach in a non-IID setting and emphasizes the need for novel methods that can handle such challenges effectively.

This section introduces FedTest, the proposed solution that addresses some challenges in the FL paradigm. FedTest leverages user-specific data for localized model training and evaluation and reinforces a robust system that prioritizes performance and privacy. 
Each learning round in the FL paradigm comprises two fundamental phases: The broadcast and collection phases. In the broadcast, the FL server broadcasts the aggregated model to the participating users. In the collection phase, users send their updated models to the FL server. 

Our proposed model structure involves the interaction between a central server and multiple users, all of whom can be trainers, while a subset can act as both trainers and testers.  The process begins when each user trains their individual models. After training, these models are sent to both the server and the tester users. The testers have an additional role, which is to calculate the received models' accuracies using their local datasets. Upon completion of their own model's training, these testers will attach the calculated accuracies of the models they have tested alongside their own model and then send them to the server. This establishes a comprehensive and reliable model evaluation and improvement system in a federated learning environment.
%In fedtest, each 
%Detailed through a comprehensive flowchart and algorithm, we unfold the intricacies of FedTest, revealing its potential to substantially enhance FL frameworks.

\begin{figure}
    \centering
    \includegraphics[width=0.48\textwidth]{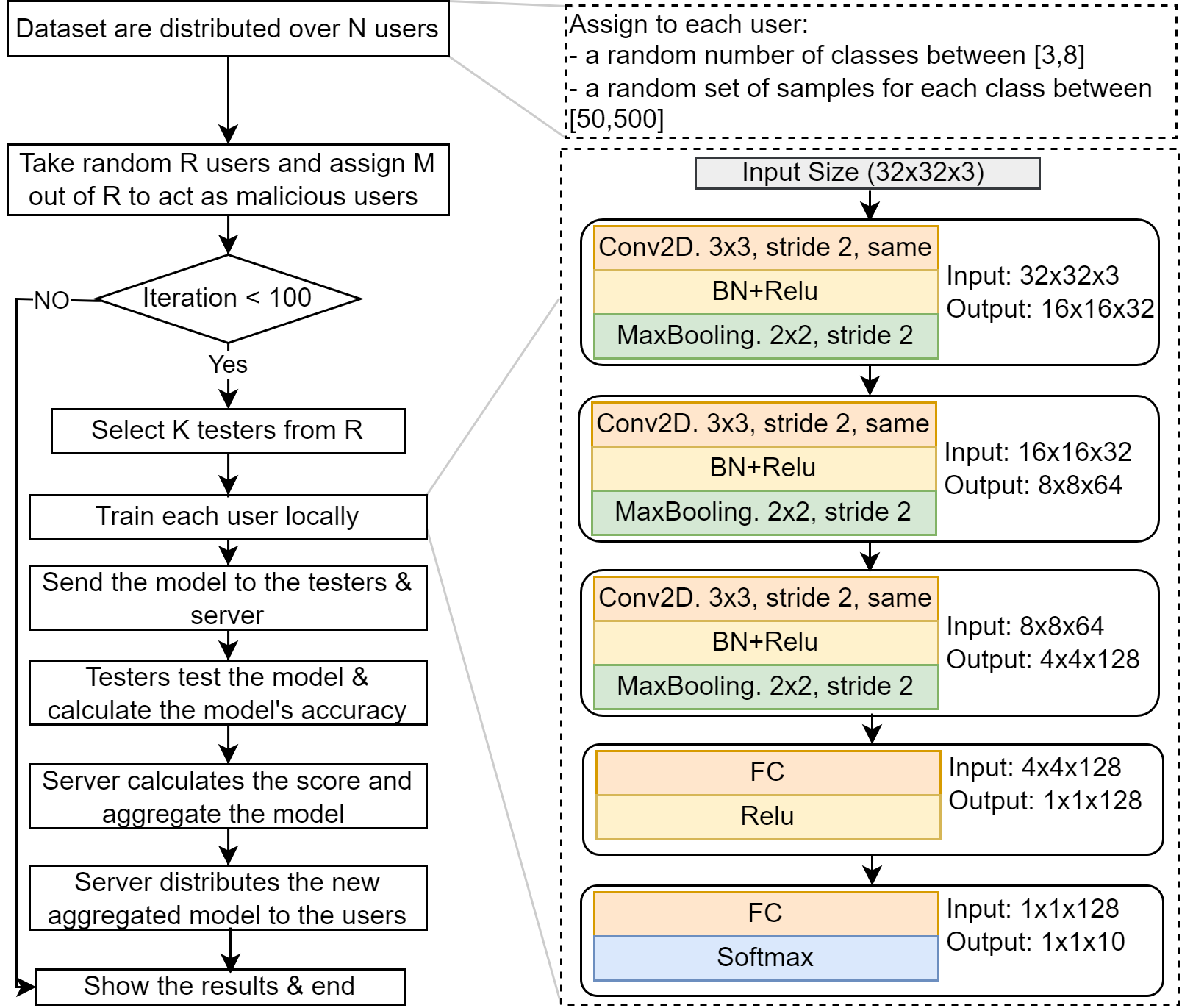}
    \caption{Flowchart of the proposed model, considering CIFAR-10 dataset.}
    \label{testchart}
\end{figure}

The proposed FedTest model operates through sequential and interactive procedures detailed in the accompanying flow chart in Fig. \ref{testchart}. Initially, we assume that the datasets are distributed among $N$ users, with each user randomly assigned a number of classes and a set of samples for each class, ensuring a non-IID data distribution. Then, a random set of users $R$ is selected (in our simulation, we assumed that $N=R$) as the users involved in training. These $R$ users train locally on their data with a model comprising three convolutional layers and two fully connected layers, followed by a softmax function, which is ideally suited for an image classification problem. We assume that of $R$ users, a number $M$ of users act as malicious users. The training process runs for several iterations, and each iteration follows a specified routine. Within each iteration, $K$ testers are selected to use their datasets to test the other models of the users.  

The users in each iteration send their models to the testers and the server. The testers evaluate the models and calculate their respective accuracies. Using their allocated resource block, the testers send their models and the calculated accuracies to the server. The server, in turn, computes scores for each model and uses them to obtain a weighted aggregated model. The scores are found based on the accuracy received in the current and previous rounds. In general, we propose using a weighted moving average approach to calculate the scores, where the recent accuracies are weighted more than the old ones. In our simulation, we noticed that the calculated scores are better if the power is increased $4$. This weighting method emphasizes the well-performing models and penalizes the poor ones. The server then broadcasts the aggregated model to all users. This process repeats until a maximum number of iterations is reached, after which the final results are displayed.

During the collection phase, we operate under the assumption that each user is allocated an orthogonal resource block (RB), which could manifest as time slots, subcarriers, or a combination of both. We assume that the testers send their models in different time slots than the other users. We consider orthogonal resource assignment to eliminate interference between users effectively. The distinct utilization of RBs enables the tester users to simultaneously receive models from other users, paralleling the server's receipt of these models. This arrangement ensures efficient use of resources and synchronizes the evaluation process between the server and testers, optimizing the system's overall operation and the potential for timely feedback.
%At the collection phase, we assume that each user utilizes an orthogonal resource (RB) block to avoid interference. These RBs can be time slots, subcarriers, or combining of both. 

%This assumption allows the testers to be able to receive other users' models at the same time the server receives these models. 
%When the testers send their models to the servers, they attach the calculated accuracy of other users' models to be utilized by the server for aggregation.
To have a clear understanding of the operations of the proposed scheme, we provide Algorithm \ref{algo1}. 

\begin{algorithm}
\caption{FedTest Algorithm}
\label{algo1}
\begin{algorithmic}[1]
\State \textbf{Inputs:} $N$ users, $K$ testing users, $M$ malicious users
\State \textbf{Initialize:} Assign each user  RB
\State Set the maximum number of iterations as $n$
\For {$iteration = 1$ to $n$} 
    \State All users train the model using their local dataset
    \For {$i=1$ to $N-K$}
        \State User $i$ sends its model at RB$_{i}$
        \State Testers conduct testing for each model in the first $N-K$ time slots
    \EndFor
    \For {$j=N-K+1$ to $N$}
        \State The tester $j$ sends its updated model and the calculated accuracies to the FL server at RB$_{j}$
    \EndFor
    \State FL server calculates the scores
    \State FL server aggregates the models using the updated scores
    \State FL server broadcasts the aggregated model to all nodes
    \State Select different $K$ testers out of $N$ nodes
\EndFor
\end{algorithmic}
\end{algorithm}

%\textcolor{red}{N users ,K testing users, Let's assume that each user is assigned with a RB where it can send its trained model.....All the users train the model using their local dataset.user N-K sends its model at $RB_{N-K}$,At the first N-K time slots, the testers conduct testing for each model.Testers send their model and calculated accuracy to the FL server at $RB_{N-K+1}$ up to $RB_{N}$.FL server calculates the scores and aggregates the models using the updated scores.FL server broadcasts the aggregated model to all nodes.Select different K testers out of N nodes.Repeat the previous points.}

%\subsection{Accuracy-based aggregation of local models}

% \begin{figure}
%     \centering
%     \includegraphics[width=0.5\textwidth]{Accuracy-FL.png}
%     \caption{Accuracy-based aggregation of local models.}
%     \label{accuracy}
% \end{figure}

% %\subsection{Federated Testing}

% \begin{figure*}
%     \centering
%     \includegraphics[width=0.6\textwidth]{newFTFL.png}
%     \caption{FedTest architecture.}
%     \label{acctesting}
% \end{figure*}

\begin{figure*}[ht]
    \centering
    \begin{subfigure}[b]{0.5\textwidth}
        \centering
        \includegraphics[width=\textwidth]{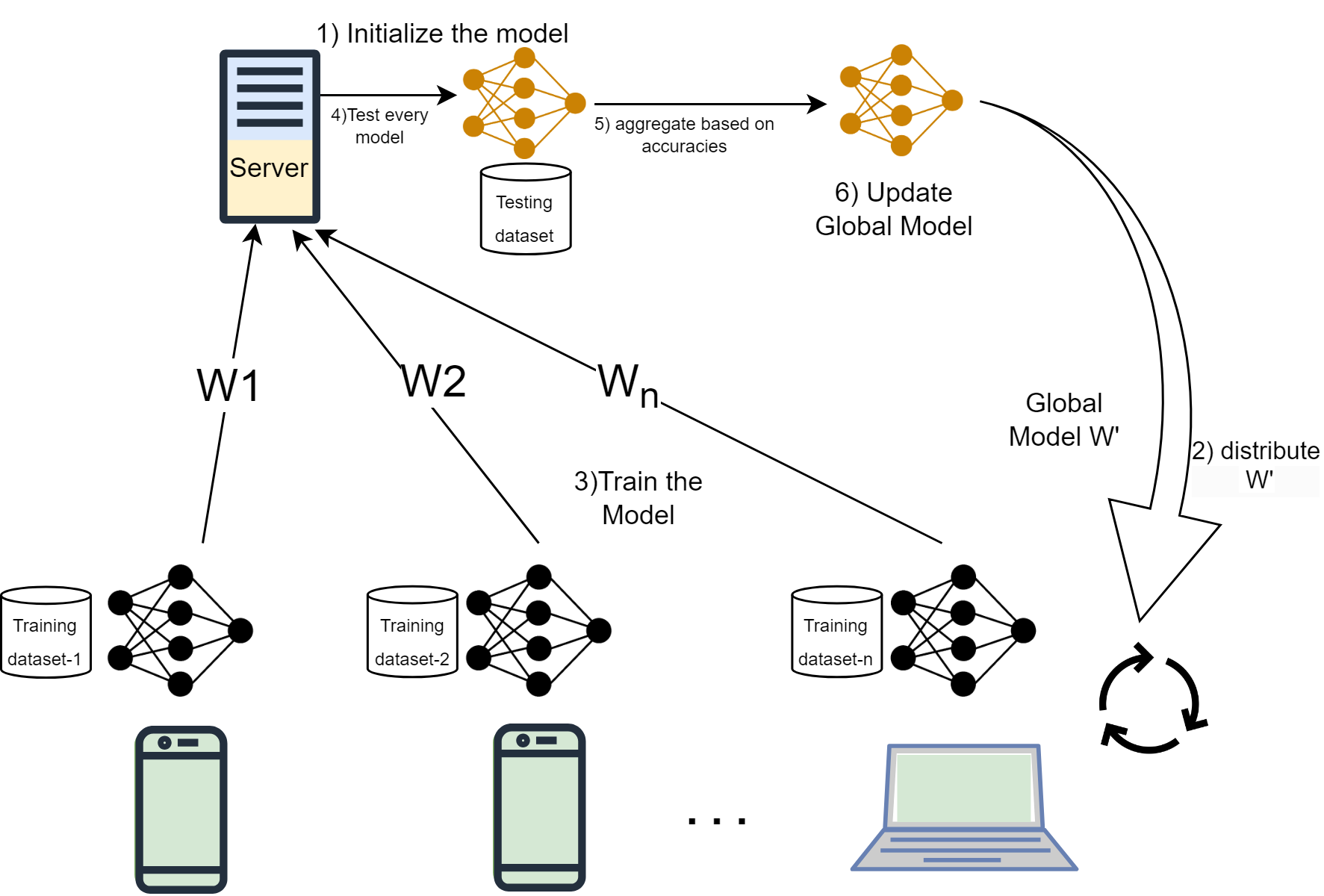}
        \caption{Accuracy-based operations cycle.}
        \label{accuracy}
    \end{subfigure}%
    \begin{subfigure}[b]{0.5\textwidth}
        \centering
        \includegraphics[width=\textwidth]{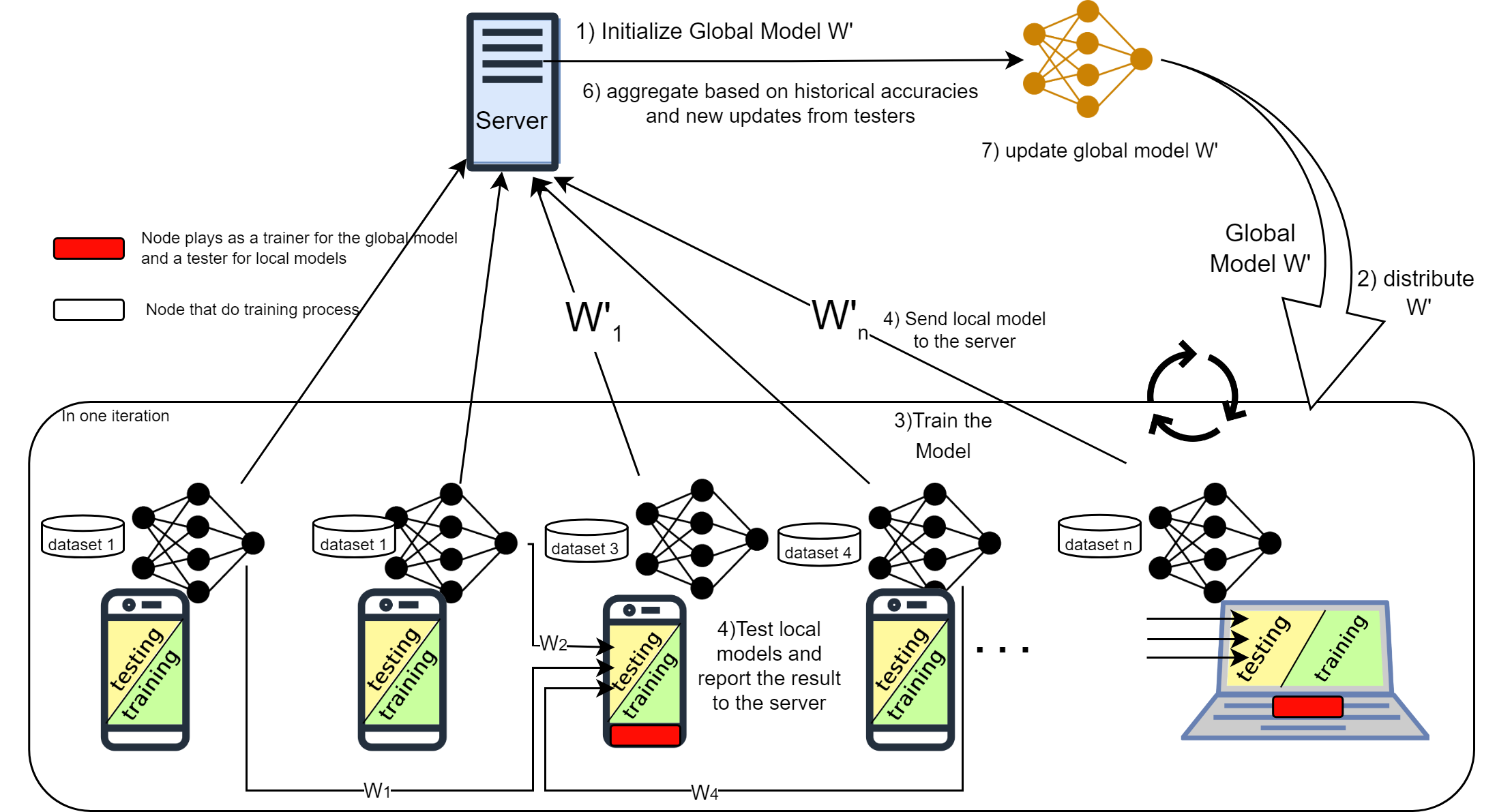}
        \caption{FedTest operations cycle.}
        \label{acctesting}
    \end{subfigure}
    \caption{The comparison between the FedTest and the accuracy-based procedures.}
    \label{fig:combined}
\end{figure*}

The strengths of the proposed FedTest model can be succinctly condensed into three pivotal points:
\begin{itemize}
    \item \textbf{Optimized Utilization and Efficiency:} FedTest model demonstrates operational efficiency by effectively leveraging user-generated data for both training and testing, while concurrently optimizing system resources. The orthogonal resource block allocation for each user prevents interference, which further underlines FedTest's commitment to optimized utilization of resources.
    \item \textbf{Mitigates Impact of Malicious Users and Handles Non-IID Data:} With an accuracy-based model evaluation and aggregation process, FedTest not only mitigates the influence of malicious users, but also handles non-IID and imbalanced data efficiently. The model actively promotes fair learning and maintains its efficacy despite uneven data distributions, thus ensuring robustness and security.
    \item \textbf{Promotes Faster Convergence Rates and Improved Performance:} By addressing the common issue of client drift in federated learning through a weighted aggregation scheme, FedTest accelerates the convergence of federated learning systems and enhances the overall performance. This innovative approach ensures that the global model stays close to optimal, contributing to improved performance and faster convergence rates.
\end{itemize}

Figures \ref{accuracy} and \ref{acctesting} show the operations cycle in the accuracy-based and FedTest, respectively. While the accuracy-based approach uses the dataset available at the FL server to test the users' model, our approach uses the distributed datasets over the users to test the models and build up accurate scores for the models. 

\section{Results and Discussion}

In this section, we examine the proposed approach using the CIFAR-10 and MNIST datasets and compare its performance to the FedAvg and accuracy-based approaches. In the figures, we show the convergence rate versus the number of global iterations for the proposed scheme and the adopted baseline schemes. We assume that the number of clients is $20$ in a federated learning environment.
%We conducted experiments with 20 selected clients in a federated learning setting, where a central server aggregates the global model. 
The experiments are implemented under two scenarios: 1) No malicious users and 2) With malicious users, where some users send random weights to the server.

\begin{figure}
    \centering
    \includegraphics[width=0.5\textwidth]{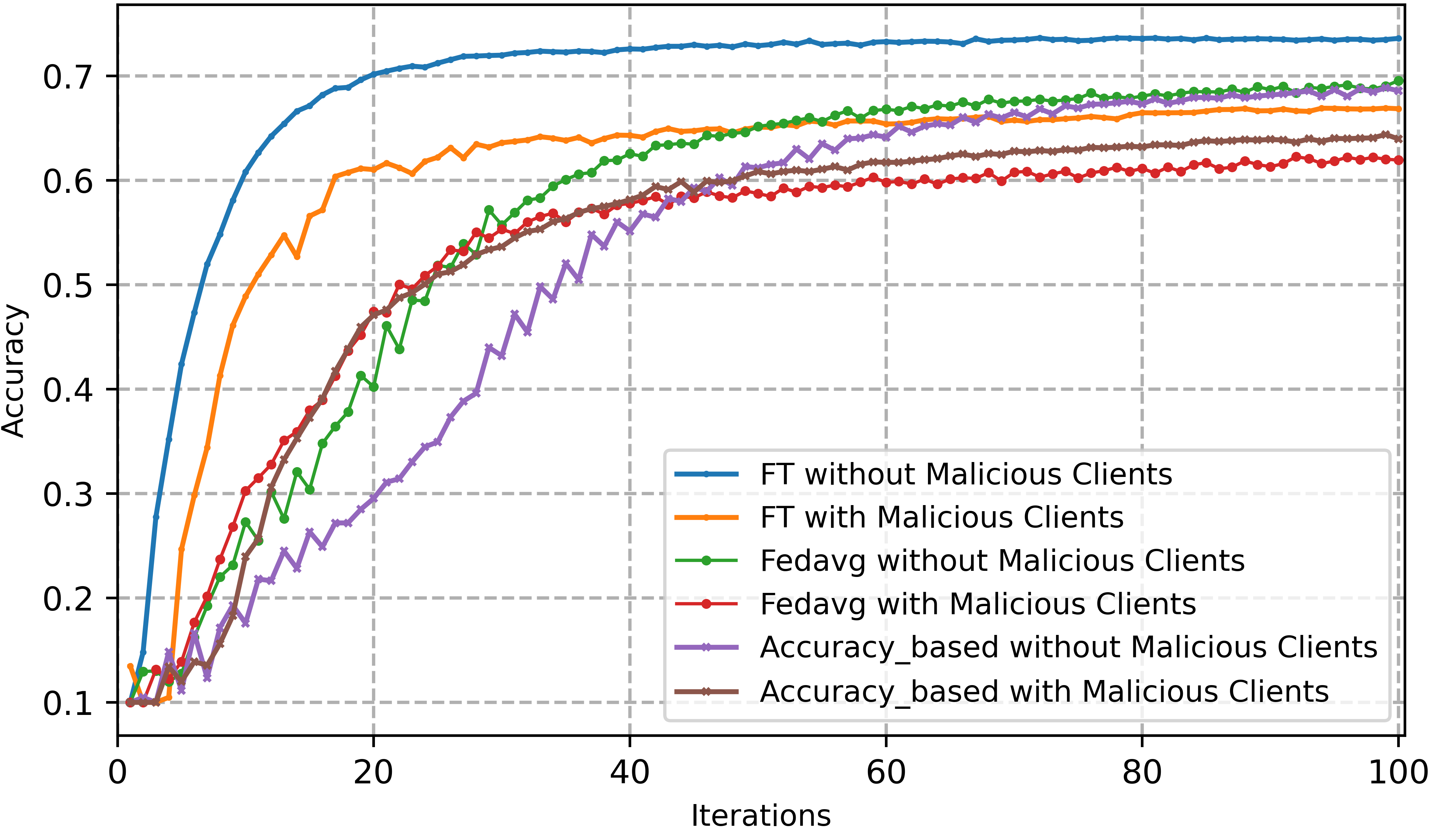}
    \caption{Convergence rate of the proposed FedTest compared to FedAvg and accuracy-based approaches using CIFAR-10 dataset.}
    \label{cifar}
\end{figure}

Fig. \ref{cifar} shows the convergence rates of the proposed FedTest, FedAvg, and accuracy-based approaches when considering the CIFAR-10 dataset. The convergence rates are plotted for both scenarios: when there are no malicious users and when 3 malicious users are present. It can be seen that the proposed approach outperforms the other methods, achieving higher accuracy and faster convergence rates in both scenarios. Specifically, in the scenario without malicious users, FedTest achieves the same accuracy in only 20 iterations compared to 100 iterations required by the other approaches. Generally, the presence of malicious users slows the convergence rate and prevents the achievement of high accuracy values. However, the proposed FedTest demonstrates faster convergence (by approximately five times) and achieves higher accuracy compared to FedAvg and the accuracy-based approach.

\begin{figure}
    \centering
    \includegraphics[width=0.5\textwidth]{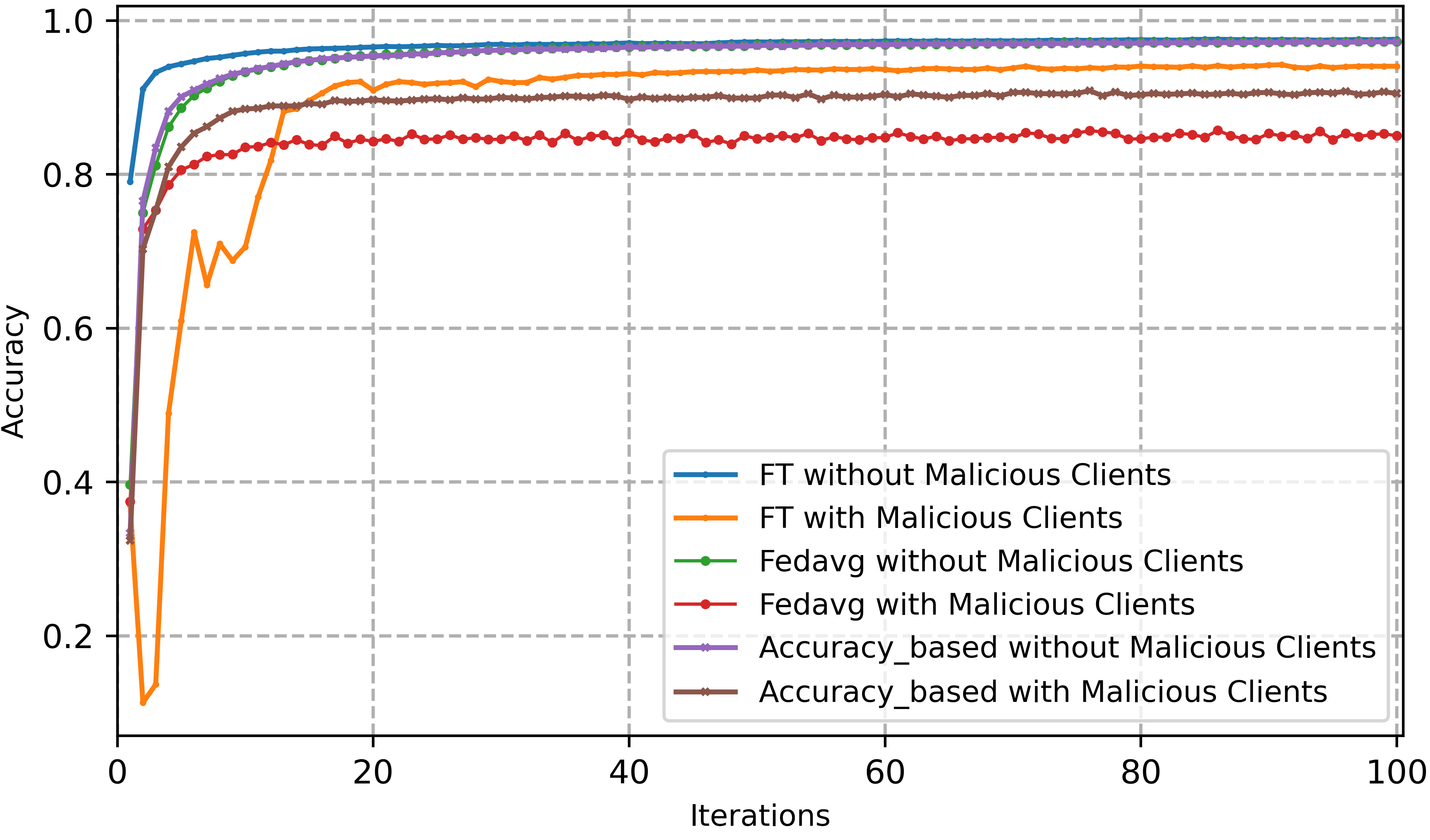}
    \caption{Convergence rate of the  proposed FedTest compared to FedAvg and accuracy-based approaches using MNIST dataset.}
    \label{mnist}
\end{figure}

Fig. \ref{mnist} shows the convergence rates for the MNIST dataset. In the scenario without malicious users, the proposed FedTest approach performs slightly better than other approaches. The figure indicates that all approaches exhibit similar behavior, with no significant differences in accuracy or convergence rates. This is because the MNIST dataset does not sufficiently challenge differentiating between strong and weak neural network models. In other words, both basic and complex models can achieve high classification accuracy in the MNIST dataset \cite{zhang2021dive, shen2019learning, xiao2017fashion, wang2022oracle}.

However, in the presence of malicious users, the classification of the MNIST dataset becomes more challenging. Thus, we evaluated the approaches with 4 malicious users. Fig. \ref{mnist} shows a significant improvement in accuracy, with FedTest outperforming other approaches in the presence of malicious clients. The figure also illustrates that FedAvg achieves the lowest accuracy, as it fails to detect malicious users and appropriately weigh their contributions.

% In the first scenario, without malicious clients, our proposed method's performance varied depending on the used dataset. For the CIFAR-10 dataset, the proposed approach outperformed the other methods, achieving higher accuracy and faster convergence rates. However, when using the MNIST dataset, all approaches displayed approximately the same behavior, with no significant difference in their accuracy or convergence rates. Although MNIST was once a reliable benchmark dataset, even basic models by current standards can attain high classification accuracy, rendering it inadequate for differentiating between more powerful and weak models \cite{zhang2021dive,shen2019learning,xiao2017fashion,wang2022oracle}. This suggests that the superiority of our proposed method is more pronounced in specific datasets, such as the CIFAR-10 dataset, while it converges to similar performance as the other approaches in other cases, like the MNIST dataset. In the second scenario, we introduced malicious clients, with 3 malicious clients in the CIFAR-10 dataset and 4 malicious clients in the MNIST dataset, to study the behavior of the approaches under varying adversarial conditions. Our proposed method demonstrates significant improvement in accuracy, outperforming other approaches even when using the MNIST dataset in the presence of malicious clients. It achieves the highest accuracy and fastest convergence, followed by the second approach, which is accuracy-based. The third approach, FedAvg, exhibited the lowest accuracy in this case. 
Overall, the figures demonstrate the robustness and efficiency of our proposed method in both scenarios and datasets, highlighting its superiority in terms of accuracy and convergence rate compared to the other two approaches.

%mnist dataset can not distinguish the powerful model from the weak one in terms of accuracy \cite{} 

\section{Research Directions}

\subsection{Optimizing the Communication Cost of FedTest}

This paper proposes the FedTest and studies its implementation from the learning perspective. However, we need to optimize the proposed FedTest to minimize the communication cost. This can be done by studying the problem of tester selection at each round, user scheduling, and RBs allocation. In particular, when each user sends its model the received rate is determined by the minimum rate between the transmitter and the BS and between the transmitter and all the testers.
Hence, the rate of tranmission depends on the testers channels. This means that selecting appropriate testers in each round would improve the transmission rates of the models. 
\subsection{Optimizing the Accuracy Score of FedTest}

In an effort to optimize the accuracy score, the implementation of our proposed method involved elevating the accuracy to the fourth power. This strategy was employed to amplify instances of high accuracy while mitigating the impact of low accuracy, which often arises due to malicious users. To further optimize the accuracy score, it is suggested that the exponent be treated as a variable, subject to periodic adjustments based on performance evaluations.

\subsection{Enhancing Models Evaluation by Identifying Malicious Users and Relying on Trusted Testers}
Engaging all users as testers within the evaluation process is unnecessary. For instance, acquiring an extensive dataset for testing purposes is superfluous if a testing dataset necessitates only 10\% of the dataset. Consequently, frequent transitions between users are not needed; rather, it is feasible to rely on a subset of more trusted users for model testing. The proposed model exhibits resilience against malicious users submitting deceptive scores. As the proposed model incorporates historical data and considers the entirety of the users, it is not exclusively reliant on a mere two or three users. Thus, the influence of false scores originating from malicious users on the overall performance is negligible. Future research could focus on devising a method to identify users who submit counterfeit or random models, thereby eliminating dependence on them during the testing phase. Through multiple rounds of global iterations, it becomes possible to discern malicious users or suspect certain users of malicious intent. Subsequently, preventative measures can be employed to limit their contributions to the model.

\subsection{Mitigation of Adversarial Attack}
FedTest can enable the system to flag models that consistently perform poorly or show suspiciously erratic behavior, which may indicate an adversarial attack. This allows the system to not only limit the influence of potentially harmful models but also potentially isolate and investigate them. The effectiveness of FedTest in mitigating adversarial attacks lies in its ability to assess and validate the quality of local models, ensuring that the aggregated global model is not unduly influenced by subpar or malicious models. This can result in a robust federated learning system that can withstand adversarial attempts to compromise its integrity.

\section{Conclusion}
This paper introduced a novel framework known as FedTest designed to amend the significant challenges of traditional FL systems, including model quality assessment, handling of unbalanced models, and the mitigation of the influence of harmful models. Our proposed method allows users to train local models and test other users' models using their own data, eliminating the need for the server to access a large testing dataset. We have demonstrated that FedTest offers clear advantages over other methodologies. It not only enables faster convergence rates, but also notably mitigates the influence of malicious users. This greatly improves the overall efficiency and robustness of FL systems, even in less robust datasets like MNIST. Our method provides a more secure and reliable learning environment, which is particularly crucial in the presence of potentially malicious users within the network.

%Moreover, it is significant to underscore that the proposed algorithm leverages the unique advantages of device-to-device (D2D) communication technology. This approach allows User Equipments (UEs) to exchange information independently without the necessity of a base station (BS) or an access point (AP), further enhancing the decentralization and efficiency of the model.

Through FedTest, we envision a more efficient, robust, and secure FL environment that could shape the future trajectory of distributed machine learning. Nonetheless, we acknowledge that while our approach shows promise, further research is required to optimize its performance, assess its scalability, and explore potential uses across an even broader array of applications.

% or


\appendices

\section*{Acknowledgment}
M. Ghaleb and M. Felemban would like to acknowledge the support received from the Interdisciplinary Research Center for Intelligent Secure Systems at KFUPM.

\ifCLASSOPTIONcaptionsoff
  \newpage
\fi



\bibliographystyle{IEEEtran}
\bibliography{fl.bib}

\end{document}